\PassOptionsToPackage{table}{xcolor}
\documentclass[10pt,twocolumn,letterpaper]{article}

\usepackage[pagenumbers]{cvpr}

\definecolor{cvprblue}{rgb}{0.21,0.49,0.74}
\usepackage[pagebackref,breaklinks,colorlinks,allcolors=cvprblue]{hyperref}

\usepackage{microtype}    
\usepackage{xcolor}        
\usepackage{graphicx}
\usepackage{booktabs}
\usepackage{multirow}
\usepackage{caption}
\usepackage{subcaption}

\def\eg{{\em e.g.}}
\def\ie{{\em i.e.}}

\definecolor{mygray}{gray}{0.9}
\definecolor{highlight}{RGB}{238,250,215}
\definecolor{lightblue}{RGB}{240,245,255}

\title{Thinking With Bounding Boxes: Enhancing Spatio-Temporal Video Grounding via Reinforcement Fine-Tuning}

\author{
    Xin Gu$^{1}$\thanks{Equal contributions \;\;\; $^{\dagger}$Corresponding author} \;\;\;
    Haoji Zhang$^{2 *}$ \;\;\;
    Qihang Fan$^3$ \;\;\;
    Jingxuan Niu$^2$ \;\;\;
    Zhipeng Zhang$^4$ \;\;\;\\
    Libo Zhang$^5$ \;\;\;
    Guang Chen$^1$ \;\;\;
    Fan Chen$^1$ \;\;\;
    Longyin Wen$^1$ \;\;\;
    Sijie Zhu$^{1\dagger}$\\
    $^1$ByteDance Intelligent Creation
    $^2$Tsinghua University\\
    $^3$Institute of Automation, Chinese Academy of Sciences
    $^4$ Shanghai Jiao Tong University \\
    $^5$Institute of Software, Chinese Academy of Sciences
}

\begin{document}
\maketitle

\begin{abstract}
Spatio-temporal video grounding (STVG) requires localizing a target object in untrimmed videos both temporally and spatially from natural language descriptions. Despite their strong language understanding, multimodal large language models (MLLMs) underperform on STVG due to misaligned training objectives and weak fine-grained region-word alignment in standard visual encoders. To address this, we propose STVG-o1, the first framework that enables off-the-shelf MLLMs to achieve state-of-the-art STVG performance without any architectural modifications. Our method introduces a bounding-box chain-of-thought mechanism that explicitly reasons about spatio-temporal locations in an intermediate step before producing the final prediction. We further design a multi-dimensional reinforcement reward function consisting of format, consistency, temporal, spatial, and think rewards, which provides geometry-aware supervision through reinforcement fine-tuning. Evaluated on HCSTVG-v1/v2 and VidSTG, STVG-o1 sets new state-of-the-art results on HCSTVG, outperforming the best task-specific method by 7.3\% m\_tIoU on HCSTVG-v1, matching specialized models on VidSTG, and surpassing all existing MLLM-based approaches by large margins. It also demonstrates strong open-vocabulary generalization across datasets, establishing MLLMs as viable and powerful backbones for precise spatio-temporal grounding. Our code and models will be released.
\end{abstract}

\section{Introduction}

\noindent
Spatio-Temporal Video Grounding (STVG) aims to localize a target object in an untrimmed video both temporally (\ie, its start and end timestamps) and spatially (\ie, bounding boxes in each frame) given a free-form natural language description~\cite{STGRN}. As a core multimodal understanding task, STVG requires fine-grained alignment between complex spatio-temporal video dynamics and textual semantics. Due to its fundamental role in video-language understanding and wide-ranging applications such as content-based video retrieval, intelligent surveillance, human-computer interaction, and service robotics, STVG has emerged as a major research focus in recent years.

\begin{figure}[!t]
    \centering
    \includegraphics[width=1.0\linewidth]{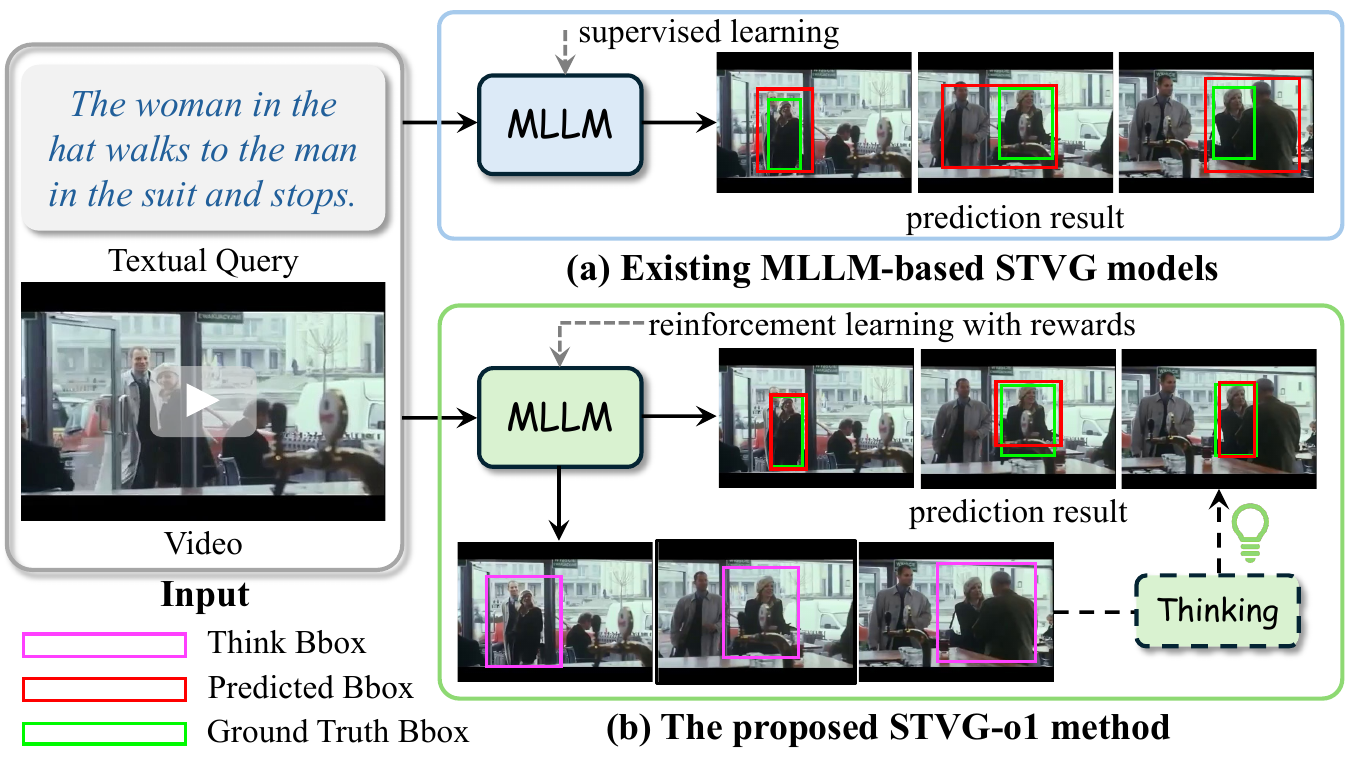}
    \caption{Comparison between existing MLLM-based STVG methods in (a) and the proposed STVG-o1 method in (b). \emph{Best viewed in color for all figures}.}
    \label{fig:1}
    \vspace{-3mm}
\end{figure}

\noindent
Existing STVG methods~\cite{STCAT, TubeDETR, cgstvg, tastvg} primarily adopt transformer-based encoder-decoder architectures that model spatio-temporal alignment between video frames and textual descriptions in an end-to-end manner, achieving strong performance. These approaches typically build upon pre-trained image-level grounding models (\eg, MDETR~\cite{mdetr} or Grounding DINO~\cite{liu2024grounding}) and transfer their spatial reasoning capabilities to videos via fine-tuning. While effective, their performance is ultimately bounded by the limited language understanding and generalization capacity of the underlying grounding backbone. This limitation becomes especially pronounced with complex or abstract queries such as ``\emph{a person wearing red clothes jumps and then exits from the left}'', where accurate semantic parsing and precise localization remain challenging. In contrast, Multimodal Large Language Models (MLLMs)~\cite{li2025llava, achiam2023gpt, comanici2025gemini, bai2025qwen2} trained on large-scale multimodal data offer significantly stronger language comprehension and cross-modal reasoning, suggesting a promising alternative for STVG. Recent works~\cite{bai2025qwen2, zhang2024video, li2024groundinggpt, wang2025spacevllm} have indeed attempted to explore such MLLM-based solutions. Despite promising results, their performance still lags far behind task-specific methods. We argue that this under-performance stems not from a lack of inherent capability but from two key mismatches. \textit{\textbf{First}}, the training objective is ill-suited for grounding: most methods generate timestamps or bounding boxes as text tokens and optimize them with cross-entropy loss, which penalizes semantically correct predictions such as ``1--9s'' versus ``0--8s'' due to minor lexical differences. As a result, the training signal poorly reflects the actual localization quality, such as the Intersection over Union (IoU). \textit{\textbf{Second}}, current MLLM-based approaches simply inherit visual encoders pretrained for global image-text alignment (\eg, CLIP-style~\cite{CLIP}), which capture holistic semantics but lack fine-grained region-word correspondence—a capability essential for aligning phrases like ``\emph{person in red}'' or ``\emph{jumping action}'' with specific spatio-temporal regions.

To address these limitations, we propose leveraging reinforcement fine-tuning (RFT) for STVG, inspired by its success in guiding large models toward task-oriented reasoning through reward signals aligned with downstream objectives. Unlike maximum-likelihood training, RFT enables the direct optimization of metrics that reflect actual performance, such as IoU or temporal overlap, making it particularly suitable for grounding tasks. Prior work has successfully applied RFT to improve fine-grained understanding in visual question answering ~\cite{xue2025adavideorag} and referring expression comprehension~\cite{yan2025videochat, bai2025univgr1, wang2503time}, suggesting its potential for STVG. Building on this insight, we present \textbf{STVG-o1}, a novel framework that unlocks the power of MLLMs for spatio-temporal grounding via reinforcement fine-tuning, as shown in Fig.~\ref{fig:1}. Particularly, drawing inspiration from the human cognitive process of ``\emph{thinking before deciding,}'' STVG-o1 introduces a novel bounding-box chain-of-thought mechanism. Given a video and a textual query, the model first generates an intermediate reasoning step, a sequence of spatio-temporal bounding boxes denoted as \texttt{<think\_bbox>}, which explicitly captures its initial hypothesis about the target object’s location across key frames. It then refines this hypothesis into the final prediction \texttt{<pred\_bbox>}. This two-stage generation enhances fine-grained region-word alignment and provides a structured intermediate signal that can be directly supervised by reward-based feedback. We design a multi-dimensional reward model consisting of five components, including \emph{format reward}, \emph{consistency reward}, \emph{temporal reward}, \emph{spatial reward}, and \emph{think reward}. These rewards jointly encourage structurally valid outputs, coherent spatio-temporal trajectories, accurate timestamp prediction, precise bounding box localization, and meaningful intermediate reasoning in \texttt{<think\_bbox>}. Using policy gradient methods, we fine-tune the MLLM end-to-end to maximize these rewards, effectively steering the model from semantic understanding toward precise spatio-temporal localization.

To the best of our knowledge, STVG-o1 is to date the first approach that unlocks the spatio-temporal grounding potential of off-the-shelf multimodal large language models (MLLMs) without any architectural modifications, relying solely on reinforcement-based fine-tuning with bounding-box chain-of-thought reasoning and a carefully designed multi-dimensional reward function. We evaluate our method on HCSTVG-v1/v2 and VidSTG. On HCSTVG, STVG-o1 achieves state-of-the-art performance, outperforming all prior methods, including all the task-specific methods, by a clear margin (\eg, +7.3\% m\_tIoU over TA-STVG on HCSTVG-v1). On VidSTG, it matches the best task-specific models while surpassing all MLLM-based approaches. Across benchmarks, it yields dramatic gains over the MLLM base model~\cite{bai2025qwen2} (\eg, +34.7\% m\_tIoU and +25.0\% m\_vIoU on HCSTVG-v1). We also explore its open-vocabulary capability by training STVG-o1 on VidSTG and testing it on HCSTVG-v1, where it achieves strong cross-dataset performance, demonstrating robust generalization to unseen concepts and language.

In summary, the contributions of this work are as follows:
\begin{itemize}[left=0pt]
    \item We propose \textbf{STVG-o1}, the first framework that enables off-the-shelf multimodal large language models (MLLMs) to perform spatio-temporal video grounding without any architectural modifications, leveraging a novel bounding-box chain-of-thought reasoning mechanism.
    \item We design a multi-dimensional reinforcement reward function consisting of format reward, consistency reward, temporal reward, spatial reward, and think reward, which provides fine-grained supervision over both intermediate reasoning and final predictions and effectively compensates for the weak region-word alignment in standard MLLM visual encoders.
    \item We demonstrate state-of-the-art performance on HCSTVG-v1/v2, competitive results with task-specific methods on VidSTG, consistent superiority over all existing MLLM-based methods, and strong open-vocabulary generalization across datasets.
\end{itemize}

\begin{figure*}[!t]
    \centering
    \includegraphics[width=0.98\linewidth]{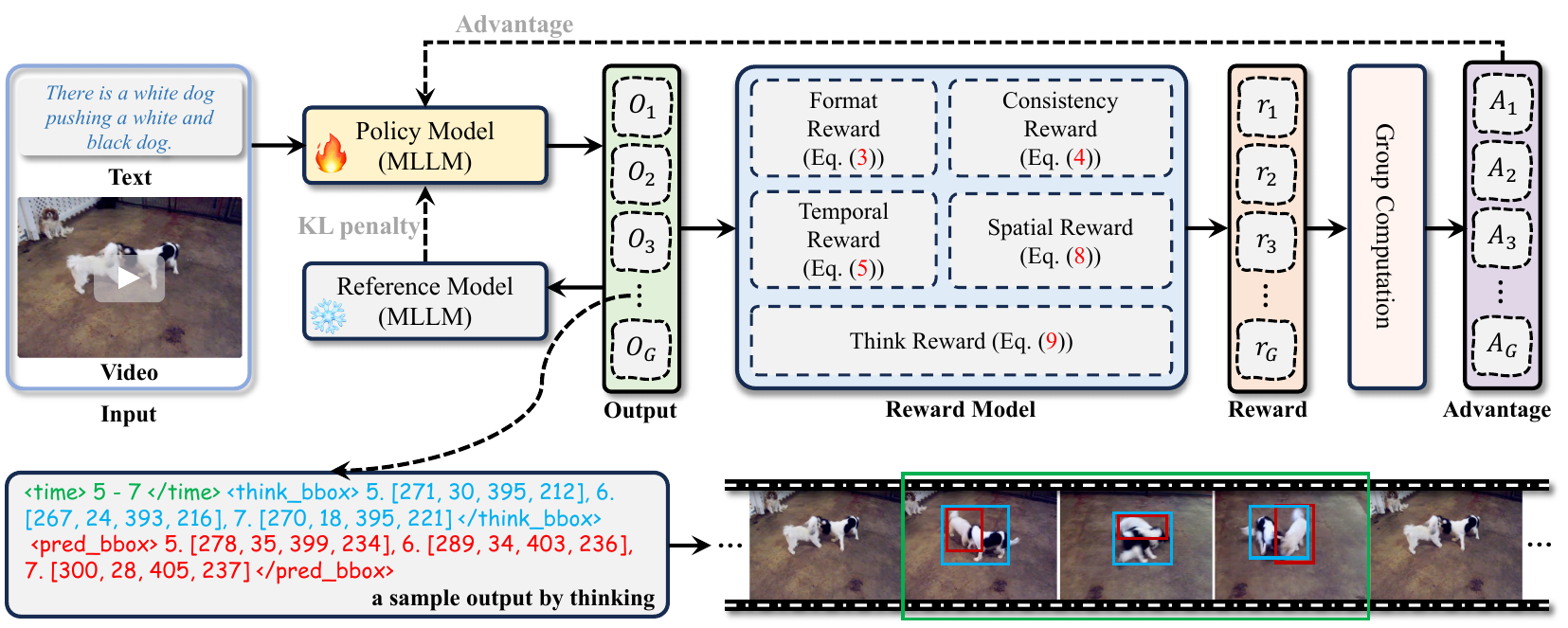}
    \caption{Overview of STVG-o1. Given a video and a natural language query, the base MLLM generates a chain-of-thought output sequence \(O_1, \dots, O_G\), where each step contains a temporal span, a sequence of thinking bounding boxes, and a sequence of final prediction bounding boxes. A reward model computes multi-dimensional rewards: format reward $\mathcal{R}_f$, consistency reward $\mathcal{R}_c$, temporal reward $\mathcal{R}_t$, spatial reward $\mathcal{R}_s$(combining GIoU and L1), and a think reward $\mathcal{R}_k$ that encourages refinement based on intermediate predictions. These rewards are aggregated to form a composite reward for reinforcement fine-tuning, enabling accurate spatio-temporal video grounding without architectural modifications. \emph{Best viewed in color for all figures}.}
    \label{fig:framework}
\end{figure*}

\section{Related Work}
\textbf{Spatio-Temporal Video Grounding.}
STVG aims to localize a target of interest in an untrimmed video based on a natural language description, requiring both temporal and spatial localization. The development of STVG methods~\cite{STGRN, OAMBRN,STCAT,TubeDETR, cgstvg, csdvl} can be broadly divided into two phases. Early approaches~\cite{STGRN, OAMBRN,STVGBert} typically followed a two-stage pipeline: they first generated candidate proposals using a pre-trained object detector and then selected the correct proposal based on the textual query. More recent methods~\cite{STCAT, cgstvg, csdvl, yao2025omnistvg} have shifted toward a single-stage encoder-decoder framework to eliminate the heavy reliance on external detection models. In this paradigm, the encoder fuses multimodal features from video and text, while the decoder directly predicts the spatio-temporal location of the target without any external detector, achieving strong performance. However, these task-specific models still struggle to comprehend complex semantic descriptions or reason effectively in visually cluttered scenarios—areas where multimodal large language models (MLLMs) exhibit significantly stronger capabilities. \textbf{In contrast}, we propose to tackle STVG by activating the inherent spatio-temporal grounding potential of off-the-shelf MLLMs, without any architectural modifications.

\noindent
\textbf{Grounding in MLLMs.}
Recent multimodal large language models (MLLMs)~\cite{achiam2023gpt, guo2025seed1, guo2025deepseek, zhu2025internvl3, bai2025qwen2, zeng2025glm} have demonstrated promising capabilities in visual grounding tasks. Most existing efforts~\cite{bai2025qwen2, guo2025seed1, ma2024groma, zhu2025internvl3, zhang2024llava, chen2023minigpt, wang2025ponder} focus on spatial grounding in static images, where the model is prompted to localize objects referred to in the text—typically by outputting bounding box coordinates or selecting from region proposals. For video understanding \cite{drft,mun2020local,hao2022can,wang2023protege,zhang2023text,lmmg, zhang2025thinking, team2025vidi}, a few MLLMs extend grounding to the temporal dimension, localizing events or actions described in language along the time axis, \eg, predicting start and end timestamps. However, current MLLM-based grounding models are limited to either spatial or temporal localization, but not both simultaneously.
\textbf{In contrast}, we propose activating the joint spatio-temporal grounding capability of off-the-shelf MLLMs through a bounding-box chain-of-thought reasoning mechanism, enabling precise localization in both space and time within untrimmed videos.

\noindent
\textbf{Reasoning-Enhanced MLLMs.}
Chain-of-thought (CoT) reasoning has been extended from language models to multimodal settings to improve the performance of MLLMs on complex vision-language tasks. Recent works~\cite{fang2025guided, bai2025univgr1, zhang2025flashvstream, wang2024hierarchical, zhang2025vtimecot, lei2025scalability, li2025tempsamp, yan2025videochat, gu2023text, wang2503time} enhance MLLMs with step-by-step visual or semantic rationales~\cite{wang2025vgr}, such as object relations or scene descriptions~\cite{wang2024world}, through in-context examples or intermediate token generation. Although effective, these approaches rarely support coordinate-level spatial or spatio-temporal reasoning. \textbf{In contrast}, we introduce a bounding-box chain-of-thought that enables MLLMs to explicitly reason about object locations over time, unlocking their potential for precise STVG.

\section{Method}

\textbf{Overview.}
We present STVG-o1, a reinforcement fine-tuned framework that enables off-the-shelf multimodal large language models (MLLMs) to perform precise spatio-temporal video grounding without architectural modifications. As detailed in $\S$\ref{framework}, our approach introduces a bounding box chain-of-thought mechanism: the model first generates intermediate spatio-temporal bounding box predictions as an explicit reasoning step and then produces refined final predictions. To provide geometry-aware supervision, we design a multi-dimensional reward function ($\S$\ref{reward}) consisting of format, consistency, temporal, spatial, and think rewards, which jointly evaluate both intermediate and final outputs. The entire system is trained via policy gradient-based optimization ($\S$\ref{optimization}) to maximize these rewards, directly aligning the MLLM’s behavior with localization accuracy rather than token-level likelihood.

\subsection{Framework of STVG-o1}
\label{framework}

STVG-o1 unlocks the latent spatio-temporal grounding ability of off-the-shelf multimodal large language models by structuring their output into a multi-stage bounding-box reasoning process. As illustrated in Fig.~\ref{fig:framework}, the framework explicitly separates intermediate reasoning from final prediction, enabling coherent spatio-temporal localization without any architectural modifications.
Specifically, given a video $\mathcal{V} = \{v_i\}_{i=1}^{N_v}$ and a natural language query $\mathcal{W} = \{w_i\}_{i=1}^{N_t}$, where $N_v$ is the video length and $N_t$ is the text length, the multimodal large language model (MLLM) first generates a raw output string:
\begin{equation}
    O_{\text{str}} = \text{MLLM}(\mathcal{V}, \mathcal{W})
\end{equation}
This string is then parsed via regular expression matching to extract three key structured components:
\begin{equation}
    \mathcal{T}^p,\ \mathcal{B}^t,\ \mathcal{B}^p = \text{RegexParse}(O_{\text{str}})
\end{equation}
These components consist of a temporal interval $\mathcal{T}^p = [t_s, t_e]$, wrapped within the \texttt{<time>} and \texttt{</time>} delimiters; an intermediate bounding box chain-of-thought $\mathcal{B}^t = \{b_i^{t}\}_{i=t_s}^{t_e}$, where each $b_i^{t} \in \mathbb{R}^4$ denotes the predicted bounding box in frame $i$, parameterized by its two opposite corner coordinates $(x_1, y_1, x_2, y_2)$, and enclosed within \texttt{<think\_bbox>} and \texttt{</think\_bbox>}; and a final spatial prediction $\mathcal{B}^p = \{b_i^{p}\}_{i=t_s}^{t_e}$ with the same parameterization, enclosed in \texttt{<pred\_bbox>} and \texttt{</pred\_bbox>}. Both the reasoning chain and the final prediction consist of bounding boxes aligned with the predicted time span $[t_s, t_e]$. This design enables end-to-end spatio-temporal grounding while preserving the original MLLM architecture and leveraging its native reasoning capacity.

\subsection{Multi-dimensional Reward}
\label{reward}

To enable precise spatio-temporal grounding, we introduce a multi-dimensional reward that jointly supervises intermediate and final predictions. It consists of five components: format, consistency, temporal, spatial, and think rewards, each targeting a distinct aspect of localization quality.

\noindent
\textbf{Format reward.} To ensure the model outputs results in the desired format, we expect it to wrap temporal localization within \texttt{<time>\dots</time>}, enclose the intermediate bounding box within \texttt{<think\_bbox>\dots</think\_bbox>}, and place the final prediction within \texttt{<pred\_bbox>\dots</pred\_bbox>}. We use regular expression matching to determine whether the model’s output conforms to the specified format:
\begin{equation}
\mathcal{R}_{\text{f}} = 
\begin{cases}
1, & \text{if output matches format,} \\
0, & \text{if output doesn't match format.}
\end{cases}
\end{equation}

\noindent
\textbf{Consistency Reward.}
To ensure coherent spatio-temporal reasoning, we require both the intermediate bounding box $\mathcal{B}^t$ and the final prediction $\mathcal{B}^p$ to strictly align with the predicted temporal interval $\mathcal{T}^p$, meaning that both sequences must represent bounding boxes for frames~$t_s, t_s+1, \dots, t_e$. A reward is given only if both $\mathcal{B}^t$ and $\mathcal{B}^p$ align with frames~$t_s$ through~$t_e$:
\begin{equation}
\mathcal{R}_{\text{c}} = 
\begin{cases}
1, & \text{if the output is consistent,} \\
0, & \text{If the output is not consistent.}
\end{cases}
\end{equation}

\noindent
\textbf{Temporal Reward.}
To improve temporal grounding, we design a temporal reward. Specifically, we use the Intersection over Union (IoU) between the predicted and ground truth temporal segments as the metric for this reward. This reward effectively guides the model to improve the precision of localizing the target event in time.

\begin{equation}
\mathcal{R}_{\text{t}} = \text{IoU}(\mathcal{T}^p,\, \mathcal{T}^\text{gt})
\label{eq:temporal_reward}
\end{equation}
\noindent
where $\mathcal{T}^\text{gt}$ denotes the ground truth temporal segments.

\noindent
\textbf{Spatial Reward.}
To improve fine-grained spatial grounding, the spatial reward is computed only over the temporal intersection between the predicted and ground-truth event intervals. 
The intersection of the predicted time span $\mathcal{T}^p$ and the ground truth $\mathcal{T}^\text{gt}$ is defined as 
\begin{equation}
\mathcal{T}^{\cap} = [\max(t_s, t_s^\text{gt}), \min(t_e, t_e^\text{gt})] = [t_s^{\cap}, t_e^{\cap}]
\end{equation}
\noindent
If $t_s^{\cap} > t_e^{\cap}$, the intersection is empty, and the spatial reward is set to zero. For a non-empty intersection, let $\mathcal{B} = \{b_i\}_{i=t_s^{\cap}}^{t_e^{\cap}}$ be the predicted bounding boxes (from either the reasoning chain or final prediction) and $\mathcal{B}^\text{gt} = \{b_i^\text{gt}\}_{i=t_s^{\cap}}^{t_e^{\cap}}$ the corresponding ground-truth boxes. The spatial reward is computed as:
\begin{equation}
\mathcal{R}_{\text{spa}} = \frac{1}{|\mathcal{T}^{\cap}|} \sum_{i=t_s^{\cap}}^{t_e^{\cap}} \left( \text{GIoU}(b_i, b_i^\text{gt}) - \|b_i - b_i^\text{gt}\|_1 \right)
\end{equation}
where $|\mathcal{T}^{\cap}| = t_e^{\cap} - t_s^{\cap} + 1$. The total spatial reward sums over both the intermediate reasoning and the final prediction:
\begin{equation}
\mathcal{R}_{\text{s}} = \mathcal{R}_{\text{spa}}^{t} + \mathcal{R}_{\text{spa}}^{p}
\label{eq:spatial_reward}
\end{equation}
This design ensures that spatial feedback is provided only when temporal localization is reasonably accurate, thereby coupling spatial and temporal learning in principled manner.

\noindent
\textbf{Think Reward.}
To encourage the model to refine its spatial predictions through reasoning, we introduce a think reward that measures the improvement from the intermediate bounding box to the final prediction. Specifically, let $R_{\text{spa}}^{t}$ and $R_{\text{spa}}^{p}$ denote the spatial rewards (computed over the temporal intersection as in Eq.~\eqref{eq:spatial_reward}) for the reasoning sequence (\texttt{<think\_bbox>}) and the final output (\texttt{<pred\_bbox>}), respectively. The reward is computed as:
\begin{equation}
\mathcal{R}_{\text{k}} = \text{max}(\mathcal{R}_{\text{spa}}^{p} - \mathcal{R}_{\text{spa}}^{t}
, 0)
\label{eq:think_reward}
\end{equation}
A positive $r_k$ indicates that the model successfully enhances its localization accuracy through deliberative reasoning, thereby reinforcing productive chain-of-thought behavior. The total reward is formulated as:
\begin{equation}
    \mathcal{R} = \mathcal{R}_f + \mathcal{R}_c + \mathcal{R}_t + \mathcal{R}_s + \lambda_k * \mathcal{R}_k
\label{eq:total_reward}
\end{equation}
where $\lambda_k$ is a hyperparameter that effectively balances the reward contributions.

\subsection{Optimization}
\label{optimization}
We optimize the policy model $\pi_\theta$ using GRPO~\cite{guo2025deepseek}. For each training sample consisting of a video $\mathcal{V}$ and a query $\mathcal{W}$, we generate $n = 8$ responses from the current policy $\pi_{\theta_{\text{old}}}$. Each response $o_i$ receives a scalar reward $\mathcal{R}(o_i)$ as defined in Equation~\eqref{eq:total_reward}. The advantage for each response is computed by normalizing rewards within the group:
\begin{equation}
    A_i = \frac{\mathcal{R}(o_i) - \mu}{\sigma + \delta}
\end{equation}
where $\mu$ and $\sigma$ are the mean and standard deviation of the group rewards, and $\delta = 10^{-6}$ ensure numerical stability. The policy update maximizes the following objective:

\begin{align*}
&\mathcal{J}_{\text{GRPO}}(\theta) = 
\mathbb{E}_{(\tilde{\mathcal{V}}, q) \sim \mathcal{D}} \Bigg[
    \frac{1}{n} \sum_{i=1}^{n} \Big(
        \min ( r_i(\theta) A_i, \text{clip}(r_i(\theta),\\ 
&1-\epsilon,\, 1+\epsilon) A_i ) 
        - \beta D_{\text{KL}}\left( \pi_\theta(\cdot \mid q) \,\|\, \pi_{\text{ref}}(\cdot \mid q) \right)
    \Big) 
\Bigg] \,\,\,\,\,\,(12)
\end{align*}
\noindent
where $r_i(\theta) = \pi_\theta(o_i \mid q) / \pi_{\theta_{\text{old}}}(o_i \mid q)$, $\epsilon$ limit the step size via clipping, and $\beta$ controls the strength of KL regularization against a frozen reference policy $\pi_{\text{ref}}$.

\section{Experiments}

\textbf{Implementation.} We implement STVG-o1 in Python using PyTorch, with Qwen2.5-VL-7B as the base model. Our training framework is based on verl~\cite{sheng2025hybridflow} and integrates components from vLLM~\cite{kwon2023efficient} to support efficient multi-modal sequence generation and policy optimization. The model is trained using the AdamW~\cite{kingma2014adam} optimizer with a learning rate of $1e-6$ and a weight decay of $1e-2$. We adopt Group Relative Policy Optimization~\cite{guo2025deepseek} with 8 rollouts per iteration and a global batch size of 128, achieved via gradient accumulation when necessary. Input videos are uniformly sampled at 2 frames per second. Each frame is resized so that its longer side is at most 336 pixels, preserving the original aspect ratio, resulting in a maximum resolution of 336×336. The think reward weight parameter $\lambda_k$ is set to 0.5.

\noindent
\textbf{Datasets.}
We evaluate our method on two standard spatio-temporal video grounding benchmarks:  HC-STVG~\cite{hcstvg} and VidSTG~\cite{STGRN}. HC-STVG focuses on multi-person scenes, where each untrimmed video is paired with a textual description of human attributes and actions. The original version, HCSTVG-v1, contains 5,660 videos (4,500 for training and 1,160 for testing). The extended HCSTVG-v2 includes 10,131 training, 2,000 validation, and 4,413 test samples. Since the test annotations for v2 are not public, we report results on the validation set, following prior work~\cite{STCAT, cgstvg, tastvg}. VidSTG consists of 6,924 untrimmed videos with 99,943 sentences (declarative and interrogative) referring to 80 object categories, forming 44,808 video–triplet instances. Following the standard split~\cite{STGRN}, it uses 5,563 / 618 / 743 videos for training, validation, and testing, associated with 80,684 / 8,956 / 10,303 sentences, respectively.

\noindent
\textbf{Metrics.} Following prior work~\cite{TubeDETR, STCAT, cgstvg, tastvg}, we adopt three standard metrics for spatio-temporal video grounding: m\_tIoU, m\_vIoU, and vIoU@R. 
m\_tIoU evaluates temporal grounding accuracy by averaging the Intersection-over-Union (tIoU) between predicted and ground-truth time intervals over all test samples. 
m\_vIoU assesses spatial grounding quality by computing the average 3D IoU across space and time between predicted and annotated spatio-temporal tubes. 
vIoU@R measures the percentage of test samples whose vIoU exceeds a threshold $R$ (\eg, $R = 0.3, 0.5$), reflecting performance under stricter localization criteria. For detailed metrics, please kindly refer to~\cite{TubeDETR}.

\begin{table*}[htb]
\setlength{\tabcolsep}{4.0pt}
	\centering
	\begin{minipage}{.5\textwidth}
		\centering
            \caption{Comparison on HCSTVG-v1 (\%).}
            \vspace{-5pt}
		\scalebox{0.75}{
			\begin{tabular}{rcccc}
				\rowcolor{mygray} 
				\specialrule{1.5pt}{0pt}{0pt}
				Methods & m\_tIoU & m\_vIoU & vIoU@0.3 &  vIoU@0.5  \\ 
				\hline\hline
                & \multicolumn{3}{c}{\textbf{\textit{Task-specific Methods}}} & \\
				STVGBert \textcolor{lightgray}{\scriptsize{[ICCV2021]}}~\cite{STVGBert} & - & 20.4 & 29.4 &  11.3  \\
				TubeDETR \textcolor{lightgray}{\scriptsize{[CVPR22]}}~\cite{TubeDETR} & 43.7 & 32.4 & 49.8 & 23.5 \\
				STCAT \textcolor{lightgray}{\scriptsize{[NeurIPS22]}}~\cite{STCAT} & 49.4 & 35.1 & 57.7 & 30.1 \\
                    SGFDN \textcolor{lightgray}{\scriptsize{[ACMMM23]}}~\cite{wang2023efficient} & 46.9 & 35.8 & 56.3 & 37.1 \\
				STVGFormer \textcolor{lightgray}{\scriptsize{[CVPR23]}}~\cite{csdvl} & - & 36.9 & 62.2 & 34.8 \\
                VG-DINO \textcolor{lightgray}{\scriptsize{[CVPR24]}}~\cite{wasim2024videogrounding}  & - & 38.3 & 62.5 & 36.1 \\
                CG-STVG \textcolor{lightgray}{\scriptsize{[CVPR24]}}~\cite{cgstvg} & 52.8 & 38.4 & 61.5 & 36.3 \\ 
            TA-STVG \textcolor{lightgray}{\scriptsize{[ICLR25]}}~\cite{tastvg} & 53.0 & 39.1 & 63.1 & 36.8  \\ 
                \hline
                  & \multicolumn{3}{c}{\textbf{\textit{MLLM-based Methods}}} & \\
                Gemini-2.5-Pro \textcolor{lightgray}{\scriptsize{[Arxiv25]}}~\cite{comanici2025gemini} & 55.1 & 25.9 & 39.1 & 9.9 \\ 
                GPT-4o \textcolor{lightgray}{\scriptsize{[Arxiv23]}}~\cite{achiam2023gpt} & 27.5 & 7.9 & 4.0 & 0.3 \\ 
                GroundingGPT \textcolor{lightgray}{\scriptsize{[ACL24]}}~\cite{li2024groundinggpt} & 22.2 & 16.7 & 15.0 & 4.9  \\ 
                Qwen2.5-VL \textcolor{lightgray}{\scriptsize{[Arxiv25]}}~\cite{bai2025qwen2}  & 25.6 & 19.1 & 20.2 & 12.6  \\ 
                LLaVA-Video-SFT \textcolor{lightgray}{\scriptsize{[TMLR25]}}~\cite{zhang2024video}  & 52.8 & 27.7 & 43.1 & 21.3  \\ 
                Qwen2.5-VL-SFT \textcolor{lightgray}{\scriptsize{[Arxiv25]}}~\cite{bai2025qwen2}  & 53.5 & 28.6 & 45.2 & 21.9  \\ 
                SpaceVLLM \textcolor{lightgray}{\scriptsize{[AAAI26]}}~\cite{wang2025spacevllm} & 56.9 & 39.3 & 66.6 & 36.9  \\ 
				\rowcolor{lightblue} STVG-o1 (ours) & \textbf{60.3} & \textbf{44.1} & \textbf{73.3} & \textbf{43.5} \\ 
                \specialrule{1.5pt}{0pt}{0pt}
		\end{tabular}}
		\label{tab:hcstvgv1}
	\end{minipage}%
	\hfill
	\begin{minipage}{.5\textwidth}
		\centering
            \caption{Comparison on HCSTVG-v2 (\%).}
            \vspace{-5pt}
		\scalebox{0.75}{
			\begin{tabular}{rcccc}
				\specialrule{1.5pt}{0pt}{0pt}
				\rowcolor{mygray} 
				Methods & m\_tIoU & m\_vIoU & vIoU@0.3 &  vIoU@0.5  \\ \hline\hline
                 & \multicolumn{3}{c}{\textbf{\textit{Task-specific Methods}}} & \\
				PCC \textcolor{lightgray}{\scriptsize{[arxiv22]}}~\cite{pcc} & - &  30.0 & - & -  \\ 
				2D-Tan \textcolor{lightgray}{\scriptsize{[arxiv22]}}~\cite{2d-tan}  & - & 30.4 &  50.4 & 18.8  \\
				MMN \textcolor{lightgray}{\scriptsize{[AAAI22]}}~\cite{mmn} & - & 30.3 & 49.0 & 25.6 \\
				TubeDETR \textcolor{lightgray}{\scriptsize{[CVPR22]}}~\cite{TubeDETR} & 53.9 & 36.4 & 58.8 & 30.6 \\
				STVGFormer \textcolor{lightgray}{\scriptsize{[CVPR23]}}~\cite{csdvl} & 58.1 & 38.7 & 65.5 & 33.8 \\
                VG-DINO \textcolor{lightgray}{\scriptsize{[CVPR24]}}~\cite{wasim2024videogrounding}  & - & 39.9 & 67.1 & 34.5 \\
                CG-STVG \textcolor{lightgray}{\scriptsize{[CVPR24]}}~\cite{cgstvg} & 60.0 & 39.5 & 64.5 & 36.3 \\ 
                TA-STVG \textcolor{lightgray}{\scriptsize{[ICLR25]}}~\cite{tastvg} & 60.4 & 40.2 & 65.8 & 36.7 \\
                \hline
                & \multicolumn{3}{c}{\textbf{\textit{MLLM-based Methods}}} & \\
                Gemini-2.5-Pro \textcolor{lightgray}{\scriptsize{[Arxiv25]}}~\cite{comanici2025gemini} & 60.4 & 24.5 & 34.6 & 9.6 \\ 
                GPT-4o \textcolor{lightgray}{\scriptsize{[Arxiv23]}}~\cite{achiam2023gpt} & 32.7 & 9.1 & 5.7 & 0.0 \\ 
                GroundingGPT \textcolor{lightgray}{\scriptsize{[ACL24]}}~\cite{li2024groundinggpt} & 19.6 & 14.7 & 16.6 & 3.1  \\ 
                Qwen2.5-VL \textcolor{lightgray}{\scriptsize{[TMLR25]}}~\cite{zhang2024video} & 22.9 & 13.0 & 15.6 & 6.4  \\
                LLaVA-Video-SFT \textcolor{lightgray}{\scriptsize{[TMLR25]}}~\cite{zhang2024video}  & 54.2 & 24.8 & 40.1 & 15.5  \\ 
                Qwen2.5-VL-SFT \textcolor{lightgray}{\scriptsize{[Arxiv25]}}~\cite{bai2025qwen2} & 55.3 & 26.5 & 38.6 & 20.2  \\
                SpaceVLLM \textcolor{lightgray}{\scriptsize{[AAAI26]}}~\cite{wang2025spacevllm} & 58.0 & 34.0 & 56.9 & 24.7  \\
				 \rowcolor{lightblue} STVG-o1 (ours) & \textbf{63.8} & \textbf{41.2} & \textbf{68.5} & \textbf{39.6} \\
				\specialrule{1.5pt}{0pt}{0pt}
		\end{tabular}}
		\label{tab:hcstvgv2}
	\end{minipage}
 \vspace{-5pt}
\end{table*}

\begin{table*}[htb]
\setlength{\tabcolsep}{10pt}
	\centering
        \caption{Comparison with existing state-of-the-art methods on VidSTG (\%).}
        \vspace{-5pt}
	\scalebox{0.81}{
		\begin{tabular}{rcccccccc}
			\specialrule{1.5pt}{0pt}{0pt}
			\rowcolor{mygray} 
			\cellcolor{mygray} & \multicolumn{4}{c}{ \cellcolor{mygray} Declarative Sentences} & \multicolumn{4}{c}{ \cellcolor{mygray}Interrogative Sentences} \\ 
			\rowcolor{mygray} 
			\multirow{-2}{*}{\cellcolor{mygray} Methods} & m\_tIoU & m\_vIoU & vIoU@0.3 &  vIoU@0.5  & m\_tIoU & m\_vIoU & vIoU@0.3 &  vIoU@0.5  \\
			\hline
			\hline
            & & & \multicolumn{3}{c}{\textbf{\textit{Task-specific Methods}}} & & & \\
			STGRN \textcolor{lightgray}{\scriptsize{[CVPR20]}}~\cite{STGRN}  &  48.5 &  19.8 & 25.8 & 14.6 &  47.0 & 18.3 & 21.1 & 12.8 \\
			STVGBert \textcolor{lightgray}{\scriptsize{[ICCV21]}}~\cite{STVGBert}  & - &  24.0 & 30.9 & 18.4 & - & 22.5 & 26.0 & 16.0 \\
			TubeDETR \textcolor{lightgray}{\scriptsize{[CVPR22]}}~\cite{TubeDETR} & 48.1 &  30.4 & 42.5 & 28.2 & 46.9 & 25.7 & 35.7 & 23.2 \\
			STCAT \textcolor{lightgray}{\scriptsize{[NeurIPS22]}}~\cite{STCAT} & 50.8 & 33.1 & 46.2 & 32.6 & 49.7 & 28.2 & 39.2 & 26.6  \\
               SGFDN \textcolor{lightgray}{\scriptsize{[ACMMM23]}}~\cite{wang2023efficient} & 45.1 & 28.3 & 41.7 & 29.1 & 44.8 & 25.8 & 36.9 & 23.9  \\
			STVGFormer \textcolor{lightgray}{\scriptsize{[CVPR23]}}~\cite{csdvl} & - & 33.7 & 47.2 & 32.8 & - & 28.5 & 39.9 & 26.2  \\ 
            CG-STVG \textcolor{lightgray}{\scriptsize{[CVPR24]}}~\cite{cgstvg} & 51.4 & 34.0 & 47.7 & 33.1 & 49.9 & 29.0 & 40.5 & 27.5 \\
            VG-DINO \textcolor{lightgray}{\scriptsize{[CVPR24]}}~\cite{wasim2024videogrounding}  & 52.0 & \textbf{34.7} & 48.1 & \textbf{34.0} & \textbf{50.8} & \textbf{29.9} & 41.0 & 27.6  \\
            TA-STVG \textcolor{lightgray}{\scriptsize{[ICLR25]}}~\cite{tastvg}  & 51.7 & 34.4 & 48.2 & 33.5 & 50.2 & 29.5 & \textbf{41.5} & \textbf{28.0} \\ \hline
            & & & \multicolumn{3}{c}{\textbf{\textit{MLLM-based Methods}}} & & & \\
            Gemini-2.5-Pro \textcolor{lightgray}{\scriptsize{[Arxiv25]}}~\cite{comanici2025gemini} & 49.9 & 22.5 & 33.6 & 12.0 & 45.4 & 13.7 & 17.3 & 8.1\\ 
            GPT-4o \textcolor{lightgray}{\scriptsize{[Arxiv23]}}~\cite{achiam2023gpt} & 38.3 & 9.2 & 7.1 & 1.6 & 39.8 & 6.1 & 3.5 & 0.6\\ 
            GroundingGPT \textcolor{lightgray}{\scriptsize{[ACL24]}}~\cite{li2024groundinggpt}  & 15.5 & 12.3 & 13.2 &  4.1 & 11.9 & 8.7 &  9.6 & 2.9  \\ 
            Qwen2.5-VL \textcolor{lightgray}{\scriptsize{[Arxiv25]}}~\cite{bai2025qwen2} & 16.8 & 10.9 & 14.3 & 5.4 & 13.8 &  8.5 & 11.3 & 4.4 \\ 
            LLaVA-Video-SFT \textcolor{lightgray}{\scriptsize{[TMLR25]}}~\cite{zhang2024video}  & 39.5 & 19.2 & 20.3 & 13.8 & 38.6 & 15.7 & 15.3 & 10.4 \\
            LLaVA-ST \textcolor{lightgray}{\scriptsize{[CVPR25]}}~\cite{li2025llava}  & 45.5 & 24.8 & 36.0 & 22.9 & 43.2 & 20.0 & 28.1 & 17.5 \\
            Qwen2.5-VL-SFT \textcolor{lightgray}{\scriptsize{[Arxiv25]}}~\cite{bai2025qwen2} & 41.6 & 20.3 & 26.1 & 15.4 & 40.9 &  17.1 & 17.6 & 13.9 \\ 
            SpaceVLLM \textcolor{lightgray}{\scriptsize{[AAAI26]}}~\cite{wang2025spacevllm} & 47.7 & 27.4 & 39.1 & 26.2 & 48.5 & 25.4 & 35.9 & 22.2 \\
		\rowcolor{lightblue} STVG-o1 (ours) & \textbf{52.1} & 33.5 & \textbf{48.4} & 32.0 & 50.5 & 27.9 & 39.8 & 26.0 \\
			\specialrule{1.5pt}{0pt}{0pt}
	\end{tabular}}
	\label{tab:vidstg}
	\vspace{-10pt}
\end{table*}

\begin{table}[htbp]
    \setlength{\tabcolsep}{2.pt}
    \centering
    \caption{Performance comparisons of the state-of-the-art on HCSTVG-v1 in open-vocabulary setting.}
    \vspace{-5pt}
    \scalebox{0.82}{
        \begin{tabular}{rcccccc}
            \specialrule{1.5pt}{0pt}{0pt}
            \rowcolor{mygray} 
            Method & Pre-training & m\_tIoU & m\_vIoU & vIoU@0.3 & vIoU@0.5 \\ \hline\hline
            TubeDETR~\cite{TubeDETR} & VidSTG & - & 16.8 & 22.3 & 9.2 \\
            STCAT~\cite{STCAT} & VidSTG & - & 22.6 & 32.1 & 20.8 \\
        CG-STVG~\cite{cgstvg} & VidSTG & 31.2 & 21.7 & 31.1 & 17.8  \\
        VG-DINO~\cite{wasim2024videogrounding} & VidSTG & - & 27.5 & 40.1 & \textbf{29.9} \\
        TA-STVG~\cite{tastvg} & VidSTG & 30.1 & 20.9 & 30.8 & 11.5 \\
            \rowcolor{lightblue} STVG-o1 (ours)& VidSTG & \textbf{45.9} & \textbf{32.2} & \textbf{50.8} & 25.5 \\
            \specialrule{1.5pt}{0pt}{0pt}
        \end{tabular}
    }
    \label{tab:open}
\end{table}

\subsection{State-of-the-Art Comparison}

\noindent
\textbf{HC-STVG Datasets.}
We evaluate STVG-o1 on HCSTVG-v1 and HCSTVG-v2, two challenging benchmarks for spatio-temporal video grounding. As shown in Tab.~\ref{tab:hcstvgv1} and Tab.~\ref{tab:hcstvgv2}, our method achieves state-of-the-art results, outperforming both specialized models and existing MLLM-based approaches. On HCSTVG-v1, STVG-o1 obtains 60.3\% m\_tIoU and 44.1\% m\_vIoU, surpassing the previous best method, TA-STVG, by +7.3\% and +5.0\%, respectively. Compared to the base model Qwen2.5-VL (25.6\% m\_tIoU), we achieve a +34.7\% absolute gain, demonstrating that reinforcement fine-tuning effectively unlocks the latent grounding capability of MLLMs. 

\noindent
On the larger HCSTVG-v2 benchmark, STVG-o1 further improves to 63.8\% m\_tIoU and 41.2\% m\_vIoU, establishing a new state of the art. Notably, our method achieves this without any architectural modifications, unlike prior approaches~\cite{wang2025spacevllm, li2025llava} that rely on additional detection heads or external modules. This underscores the effectiveness of our “thinking with bounding boxes” mechanism combined with task-aligned reinforcement learning rewards. The results show that off-the-shelf MLLMs, when optimized with localization-aware objectives, can match or surpass specialized STVG models.

\noindent
\textbf{VidSTG Dataset.} We also evaluate STVG-o1 on the VidSTG benchmark, which includes both declarative and interrogative language queries to test grounding robustness under diverse linguistic forms. As shown in Tab.~\ref{tab:vidstg}, our method achieves competitive performance against specialized task-specific models while consistently outperforming all existing MLLM-based approaches. On declarative sentences, STVG-o1 significantly surpasses the best prior MLLM-based method, SpaceVLLM (47.7\% m\_tIoU, 27.4\% m\_vIoU), by 4.4\% and 6.1\%, respectively. For interrogative sentences, STVG-o1 achieves 50.5\% m\_tIoU and 27.9\% m\_vIoU, again outperforming all MLLM-based methods and remaining competitive with task-specific architectures. Notably, compared to the Qwen2.5-VL-SFT, which has been supervised fine-tuned on the STVG dataset, our method yields absolute gains of 10.5\% in m\_tIoU and 13.2\% in m\_vIoU, highlighting the effectiveness of our STVG-o1.

\subsection{Open-Vocabulary Comparison}

To assess open-vocabulary generalization, we train STVG-o1 on VidSTG and evaluate it on the HCSTVG-v1 test set. As shown in Tab.~\ref{tab:open}, our method achieves 45.9\% m\_tIoU and 32.2\% m\_vIoU, substantially outperforming the TA-STVG by +15.8\% in m\_tIoU and +11.3\% in m\_vIoU. This strong cross-dataset performance highlights a key advantage of our approach: by fine-tuning a general-purpose MLLM with reinforcement learning guided by spatio-temporal rewards, STVG-o1 learns to interpret arbitrary object descriptions through language without relying on fixed detection vocabularies or task-specific modules. In contrast, most prior methods suffer from domain shift or limited linguistic coverage when evaluated outside their training distribution. The high vIoU@0.3 (50.8\%) further confirms that our predictions maintain accurate spatial alignment even under open-vocabulary conditions.

\subsection{Ablation Study}

\noindent
\textbf{Impact of thinking with bounding boxes.} We conduct ablation studies to evaluate the effectiveness of our bounding box chain-of-thought mechanism. As shown in Tab.~\ref{tab:ablation_thinking}, the base model, without any training, achieves poor performance, highlighting the need for task-specific optimization. Supervised fine-tuning (SFT) significantly improves results, but reinforcement fine-tuning (RFT), guided by our multi-dimensional reward function, further boosts performance across all metrics, demonstrating the benefit of aligning training with downstream grounding objectives through geometry-aware rewards. Most notably, adding the \texttt{<bbox\_think>} reasoning step under RFT yields substantial gains of +1.8\% m\_tIoU and +2.3\% m\_vIoU, indicating that explicit intermediate reasoning enhances both temporal and spatial localization accuracy. This confirms that the thinking-with-bounding-boxes mechanism, when supervised by our multi-dimensional rewards, effectively guides the model toward more precise spatio-temporal predictions.
\vspace{-2mm}

\begin{table}[ht]
    \centering
    \small
    \setlength{\tabcolsep}{4.5pt}
    \caption{Ablations of thinking with bounding boxes.}
    \vspace{-5pt}
    \scalebox{0.98}{
        \begin{tabular}{ccccc}
            \specialrule{1.5pt}{0pt}{0pt}
            \rowcolor{mygray} 
             Training & m\_tIoU & m\_vIoU & vIoU@0.3 & vIoU@0.5 \\ \hline\hline
            - & 25.6 & 19.1 & 20.2 & 12.6 \\ 
            SFT & 53.5 & 28.6 & 45.2 & 21.9 \\
             RFT & 58.5 & 41.8 & 68.7 & 38.9 \\
             RFT + Thinking & \textbf{60.3} & \textbf{44.1} & \textbf{73.3} & \textbf{43.5}  \\
            \specialrule{1.5pt}{0pt}{0pt}
        \end{tabular}
    }
    \label{tab:ablation_thinking}
    \vspace{-2mm}
\end{table}

\noindent
\textbf{Impact of the think reward.} We ablate the think reward, which encourages the model to refine bounding boxes based on its intermediate reasoning. As shown in Tab.~\ref{tab:thinkreward}, removing this reward leads to a drop in performance: m\_tIoU decreases from 60.3\% to 59.3\%, and m\_vIoU drops from 44.1\% to 42.2\%. This confirms that the think reward plays a key role in guiding the MLLM to iteratively improve spatial localization accuracy during reinforcement learning, enabling more precise grounding through self-corrective reasoning.
\vspace{-2mm}

\begin{table}[ht]
    \centering
    \small
    \setlength{\tabcolsep}{3.0pt}
    \caption{Ablations of think reward.}
    \vspace{-5pt}
    \scalebox{1.0}{
        \begin{tabular}{ccccc}
            \specialrule{1.5pt}{0pt}{0pt}
            \rowcolor{mygray} 
             Think Reward & m\_tIoU & m\_vIoU & vIoU@0.3 & vIoU@0.5 \\ 
            \hline\hline
             - & 59.3 & 42.2 & 69.1 & 39.9 \\
             \checkmark  & \textbf{60.3} & \textbf{44.1} & \textbf{73.3} & \textbf{43.5}  \\
            \specialrule{1.5pt}{0pt}{0pt}
        \end{tabular}
    }
    \label{tab:thinkreward}
    \vspace{-2mm}
\end{table}

\noindent
\textbf{Impact of spatial reward.} We analyze the spatial reward, which consists of GIoU and L1 distance reward components. As shown in Tab.~\ref{tab:spatialreward}, using only GIoU or L1 leads to inferior performance (\eg, m\_vIoU drops to 40.6\% and 39.9\%, respectively). In contrast, combining both achieves 60.3\% m\_tIoU and 44.1\% m\_vIoU, significantly improving spatial grounding accuracy and demonstrating the effectiveness of joint optimization.
\vspace{-2mm}

\begin{table}[ht]
    \centering
    \setlength{\tabcolsep}{3.5pt}
    \caption{Ablations of spatial reward.}
    \vspace{-5pt}
    \scalebox{0.9}{
        \begin{tabular}{ccccc}
            \specialrule{1.5pt}{0pt}{0pt}
            \rowcolor{mygray} 
             Spatial Reward & m\_tIoU & m\_vIoU & vIoU@0.3 & vIoU@0.5 \\ 
            \hline\hline
             $\mathcal{G}$ & 58.6 & 40.6 & 67.6 & 36.6 \\
             $\mathcal{L}_1$ & 58.5 & 39.9 & 67.0 & 34.6 \\
             $\mathcal{G}$ + $\mathcal{L}_1$  & \textbf{60.3} & \textbf{44.1} & \textbf{73.3} & \textbf{43.5}  \\
            \specialrule{1.5pt}{0pt}{0pt}
        \end{tabular}
    }
    \label{tab:spatialreward}
    \vspace{-2mm}
\end{table}

\noindent
\textbf{Impact of think reward weight.} We study the effect of the think reward weight $\lambda_k$ on performance. As shown in Tab.~\ref{tab:rewardweight}, we can see that the model performs best by setting $\lambda_k$ to 0.5.\vspace{-2mm}

\begin{table}[ht]
    \centering
    \setlength{\tabcolsep}{8pt}
    \caption{Ablations of think reward weight.}
    \vspace{-5pt}
    \scalebox{0.9}{
        \begin{tabular}{ccccc}
            \specialrule{1.5pt}{0pt}{0pt}
            \rowcolor{mygray} 
             $\lambda_k$ & m\_tIoU & m\_vIoU & vIoU@0.3 & vIoU@0.5 \\ 
            \hline\hline
             $0.2$ & 59.5 & 42.5 & 70.9 & 40.6 \\
             $0.5$  & 60.3 & \textbf{44.1} & \textbf{73.3} & \textbf{43.5}  \\
             $1.0$  & \textbf{60.5} & 43.8 & 72.4 & 42.9 \\
            \specialrule{1.5pt}{0pt}{0pt}
        \end{tabular}
    }
    \label{tab:rewardweight}
    \vspace{-2mm}
\end{table}

\subsection{Grounding Performance Across Object Scales}

\begin{figure}[ht]
    \centering
    \includegraphics[width=0.95\linewidth]{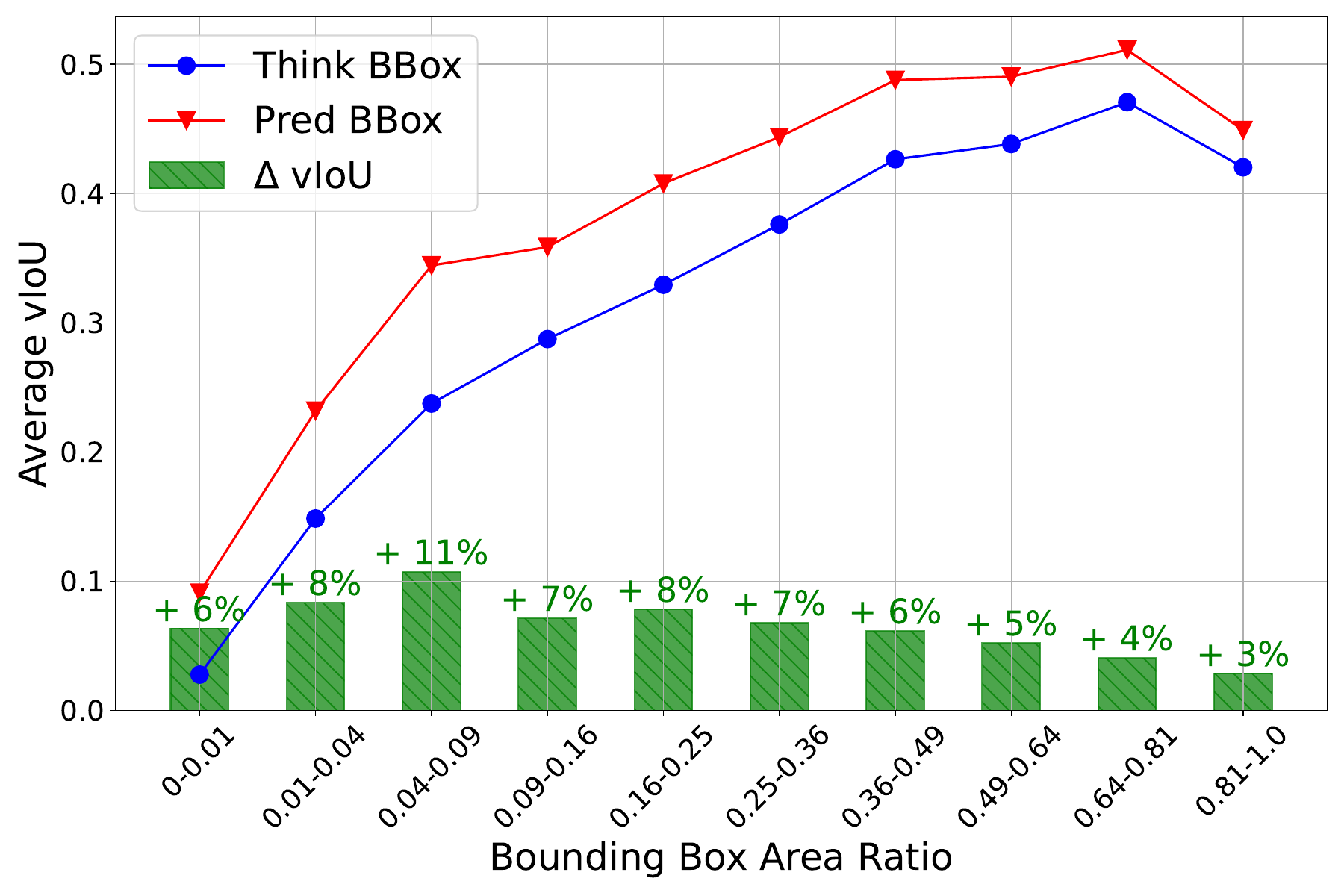}
    \caption{Analysis of average vIoU across different bounding box areas. Blue curve shows performance of intermediate \texttt{<think\_bbox>} predictions, red curve shows final \texttt{<pred\_bbox>} outputs, and green bars indicate the relative improvement ($\Delta$ vIoU) from think bbox to predicted bbox.}
    \label{fig:box_size}
    \vspace{-4mm}
\end{figure}

To understand how STVG-o1 performs spatial grounding across object scales, we analyze the average vIoU of intermediate \texttt{<think\_bbox>} and final \texttt{<pred\_bbox>} predictions, grouped by bounding box area. As shown in Fig.~\ref{fig:box_size}, the model achieves a lower vIoU in the early reasoning stage (blue curve), indicating initial coarse localization. However, the final output (red curve) consistently outperforms the intermediate prediction across all size ranges, with a maximum improvement of +11\% in the 0.04–0.09 area ratio bin, which typically corresponds to medium-sized objects. This suggests that our reinforcement-based refinement process is particularly effective for objects of moderate scale, where both visual context and language cues are rich enough to guide accurate adjustments. Notably, even for small or large objects, the \texttt{<pred\_bbox>} maintains higher accuracy than \texttt{<think\_bbox>}, confirming that the model reliably refines its spatial predictions, regardless of object size. These results validate that our method robustly enhances grounding precision across diverse object scales.

\begin{figure*}[!t]
    \centering
    \includegraphics[width=0.95\linewidth]{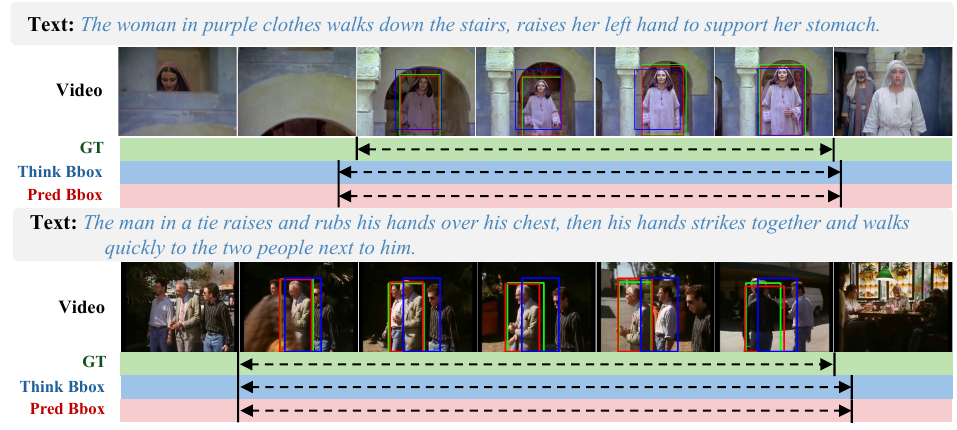}
    \vspace{-3mm}
    \caption{Qualitative results of our STVG-o1. Green boxes denote ground truth, blue boxes represent intermediate \texttt{<think\_bbox>} predictions during reasoning, and red boxes indicate final \texttt{<pred\_bbox>} outputs. \emph{Best viewed in color for all figures}.}
    \label{fig:qualitate}
    \vspace{-3mm}
\end{figure*}

\subsection{Inference Complexity Analysis}
We analyze the inference complexity of STVG-o1 and compare it with existing methods on a single GPU. As shown in Tab.~\ref{tab:complexity}, the two task-specific models, CG-STVG~\cite{cgstvg} and TA-STVG~\cite{tastvg}, achieve fast inference due to their lightweight architectures and non-autoregressive design. In contrast, all MLLM-based approaches, such as Qwen2.5-VL-SFT~\cite{bai2025qwen2}, LLaVA-ST~\cite{li2025llava}, and our STVG-o1, require autoregressive generation of bounding boxes, resulting in significantly higher latency (approximately 15–17 seconds). Please note that since SpaceVLLM~\cite{wang2025spacevllm} has not been released, we do not include it in this comparison. Despite the shared AR generation paradigm, STVG-o1 achieves faster inference than both Qwen2.5-VL-SFT (15.49 s \textit{vs.} 16.37 s) and LLaVA-ST (15.49 s \textit{vs.} 17.63 s), thanks to its more efficient and compact output format for bounding boxes. It also uses less GPU memory (24.2 GB \textit{vs.} 38.7 GB) than LLaVA-ST. These results demonstrate that while MLLMs introduce higher inference costs compared to task-specific models, they can still achieve competitive speed and better efficiency when optimized with a streamlined output design.

\begin{table}[!t]
\centering
\renewcommand{\arraystretch}{1}
\caption{Comparison on model complexity. The `AR' refers to autoregressive generation of bounding boxes.}
\scalebox{0.85}{
    \begin{tabular}{rcccc}
    \specialrule{1.5pt}{0pt}{0pt}
    \rowcolor{mygray} 
\cellcolor{mygray} Methods &  \cellcolor{mygray} AR & \cellcolor{mygray} Params & \cellcolor{mygray} Time & \cellcolor{mygray} GPU Mem \\
    \hline
    \hline
    CG-STVG~\cite{cgstvg} & & 231 M & 0.61 s & 29.7 G \\
    TA-STVG~\cite{tastvg} & & 234 M & 0.57 s & 28.4 G \\ 
    Qwen2.5-VL-SFT~\cite{bai2025qwen2} & \checkmark & 8.3 B & 16.37 s & 26.6 G \\
    LLaVA-ST~\cite{li2025llava} & \checkmark & 8.3 B & 17.63 s & 38.7 G \\ \hline
    STVG-o1 (ours) & \checkmark & 8.3 B & 15.49 s & 24.2 G \\
    \specialrule{1.5pt}{0pt}{0pt}
    \end{tabular}}
    \label{tab:complexity}
    \vspace{-3mm}
\end{table}

\subsection{Qualitative Analysis}
To further qualitatively validate the effectiveness of our proposed method,  
Fig.~\ref{fig:qualitate} shows the qualitative results of STVG-o1 on HCSTVG~\cite{hcstvg}.  
Green boxes denote ground truth, blue boxes represent intermediate \texttt{<think\_bbox>} predictions during reasoning, and red boxes indicate final \texttt{<pred\_bbox>} outputs. The model first generates a coarse localization (blue) and then refines it to better align with the target (red), illustrating how the bounding-box chain-of-thought enables progressive spatial grounding.

\noindent
Due to limited space, please refer to supplementary material for more visualizations, analyzes, and experimental details.

\section{Conclusion}

We present STVG-o1, the first framework that enables off-the-shelf multimodal large language models to perform spatio-temporal video grounding without architectural changes. By introducing a bounding-box chain-of-thought mechanism and a multi-dimensional reward function consisting of format reward, consistency reward, temporal reward, spatial reward, and think reward, we provide fine-grained, geometry-aware supervision through reinforcement fine-tuning. This design effectively overcomes the misaligned training objective and the weak region-word alignment inherent in standard MLLMs. STVG-o1 achieves state-of-the-art results on HCSTVG-v1/v2, matches specialized models on VidSTG, and outperforms all existing MLLM-based approaches. It also demonstrates strong open-vocabulary generalization across datasets. Our work shows that, with proper task-oriented rewards, MLLMs can be efficiently adapted to complex grounding tasks.

{
    % \small
    \bibliographystyle{ieeenat_fullname}
    \bibliography{main}
}

\maketitlesupplementary

\noindent
For a better understanding of this work, we offer additional details, analysis, and results as follows:

\vspace{0.3em}
\begin{itemize}
    \item \textbf{A \;\; \emph{Analysis of Qualitative Results}} \\
   In this section, we show qualitative results of our method and a comparison to the TA-STVG method.

   \item \textbf{B \;\; \emph{Analysis of Failure Cases}} \\
   In this section, we discuss the failure cases of our proposed method.

   \item \textbf{C \;\; \emph{Prompt Details for Closed-Source Model}} \\
   In this section, we present the prompt used to perform STVG with closed-source models.
   
\end{itemize}

\section*{A. Analysis of Qualitative Results}

\subsection*{A.1 Thinking with Bounding Boxes}

\begin{figure*}[!t]
    \centering
    \includegraphics[width=0.9\linewidth]{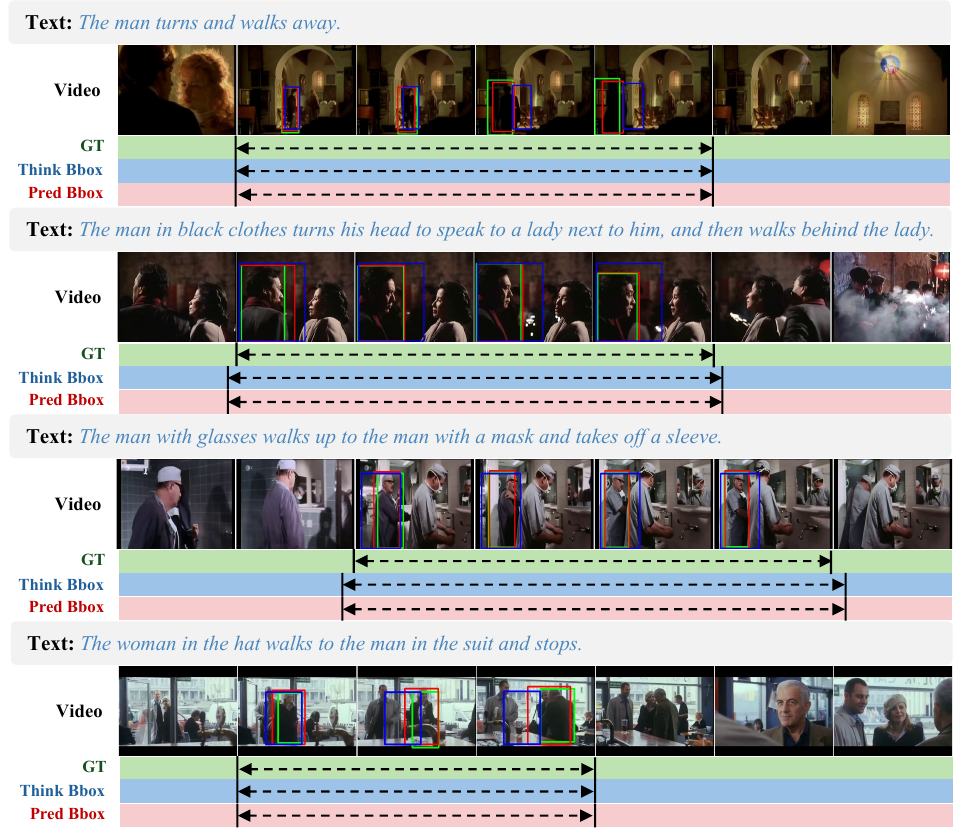}
    \caption{Qualitative results of our STVG-o1. Green boxes denote ground truth, blue boxes represent intermediate \texttt{<think\_bbox>} predictions during reasoning, and red boxes indicate final \texttt{<pred\_bbox>} outputs. \emph{Best viewed in color for all figures}.}
    \label{fig:qualitate_1}
\end{figure*}

We present qualitative results in Fig.~\ref{fig:qualitate_1} to illustrate how STVG-o1 refines its spatio-temporal predictions through a structured chain-of-thought process. In each example, the intermediate \texttt{<think\_bbox>} (blue) represents the model's initial reasoning output, while the final \texttt{<pred\_bbox>} (red) reflects the refined prediction after iterative refinement. As shown, the model consistently generates plausible but slightly imprecise bounding boxes during reasoning, often exhibiting minor localization errors or temporal drift. However, the final prediction demonstrates significant improvement; it aligns more closely with the ground truth (green), both spatially and temporally. For instance, in the first example, the initial blue box fails to fully capture the man’s movement trajectory, but the red box correctly adjusts to track his motion across frames. Similarly, in the third example, the think bbox incorrectly includes part of the background, whereas the pred bbox tightens the region to focus on the target. These observations highlight that STVG-o1 does not merely generate a single-shot prediction; instead, it engages in a deliberative reasoning process in which intermediate outputs serve as stepping stones toward accurate grounding. The progressive refinement from \texttt{<think\_bbox>} to \texttt{<pred\_bbox>} underscores the effectiveness of our reward-driven optimization in encouraging the model to improve upon its own reasoning, leading to robust and precise spatio-temporal grounding.

\subsection*{A.2 Comparison with TA-STVG}

\begin{figure*}[ht]
    \centering
    \includegraphics[width=0.9\linewidth]{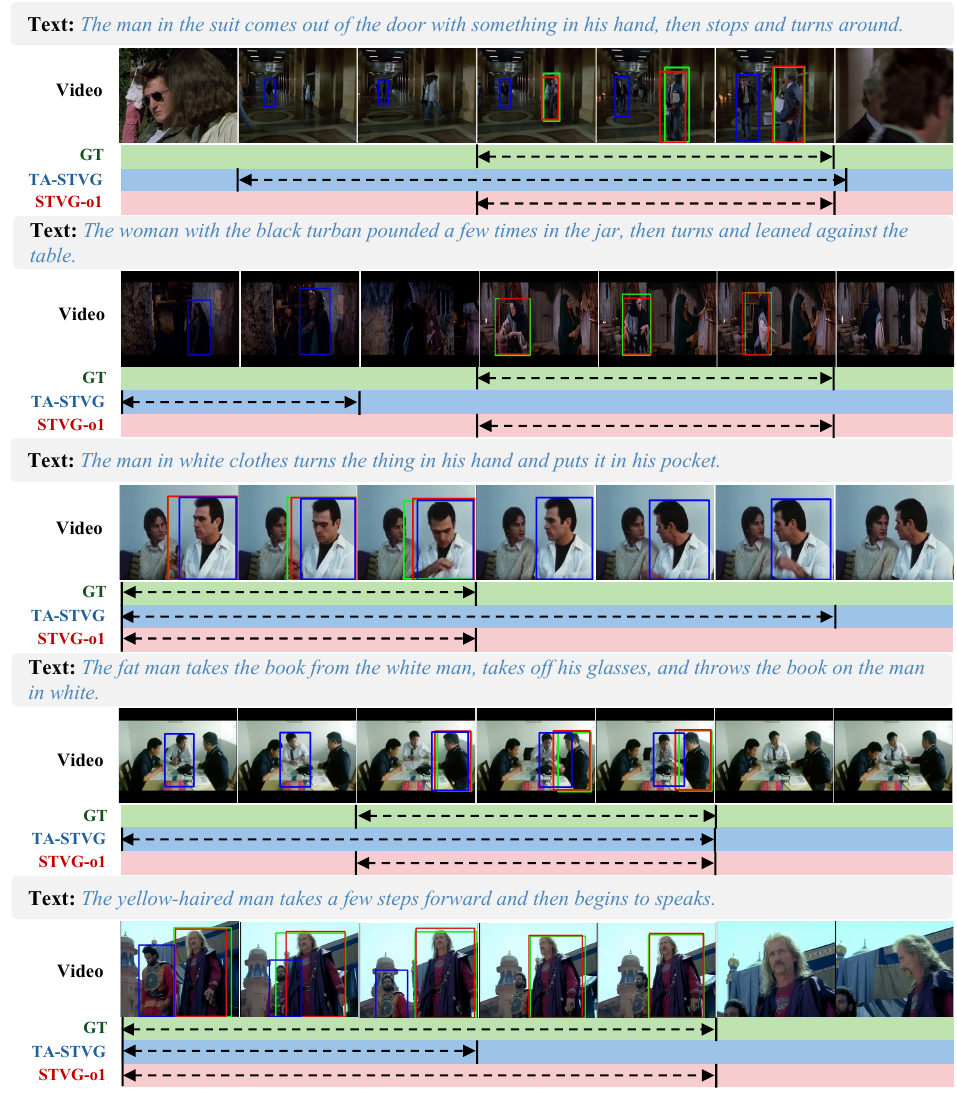}
    \caption{Qualitative results comparing our STVG-o1 with the task-specific TA-STVG method. Green boxes denote ground truth, blue boxes represent TA-STVG's predictions, and red boxes indicate our STVG-o1 outputs. \emph{Best viewed in color for all figures}.}
    \label{fig:qualitate_2}
\end{figure*}

We also present qualitative comparisons between our STVG-o1 and the task-specific method TA-STVG in Fig.~\ref{fig:qualitate_2}. Green boxes denote ground truth, blue boxes represent TA-STVG's predictions, and red boxes indicate our STVG-o1 outputs. The results demonstrate that STVG-o1 achieves more accurate and semantically consistent spatio-temporal grounding. For example, in the first instance, TA-STVG activates its prediction prematurely, starting before the target man in the suit is visible. This leads to an incorrect temporal onset that includes irrelevant motion prior to the actual event. In contrast, STVG-o1 waits until the subject becomes visually distinct, specifically when he steps fully into view, before initiating the grounding process. These results show the effectiveness of STVG-o1.

\section*{B. Analysis of Failure Cases}

\begin{figure*}[!t]
    \centering
    \includegraphics[width=0.8\linewidth]{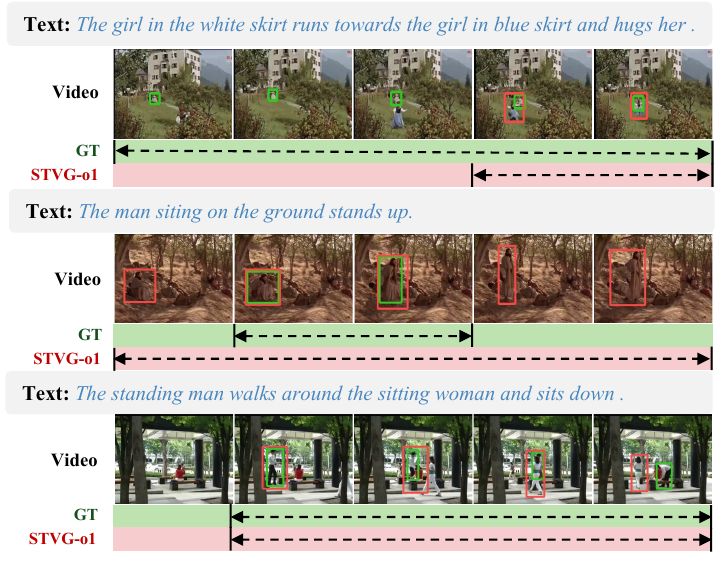}
    \caption{Failure cases of STVG-o1. Green boxes denote ground-truth bounding boxes, and red boxes indicate predictions from our method. \emph{Best viewed in color for all figures}.}
    \label{fig:fail}
\end{figure*}

Despite the strong performance of STVG-o1, it still faces challenges in certain scenarios, as illustrated in Fig.~\ref{fig:fail}. We analyze three typical failure modes: \textit{(i)} \textbf{Missing small objects.} Due to the limited resolution of input video frames, our model struggles to localize small or distant targets, especially during their initial appearance. As shown in the top example of Fig.~\ref{fig:fail}, the girl is not localized until she moves closer, which indicates a limitation in early-stage grounding.
\textit{(ii)} \textbf{Extremely short events.} With a fixed frame sampling rate, brief actions may span fewer than two sampled frames, causing critical transitions to be missed entirely. As illustrated in the middle example, the model fails to pinpoint the exact moment when the man stands up, resulting in an over-extended duration prediction.
\textit{(iii)} \textbf{Occlusion-induced confusion.} When the target object is occluded by another object, the model may shift its attention to a visually similar but incorrect entity. As seen in the bottom example (Fig.~\ref{fig:fail}), after the standing man is partially blocked by a woman, the model incorrectly localizes the woman instead, highlighting the challenge of maintaining consistent grounding under occlusion. To address these limitations, we plan to explore adaptive frame sampling, dynamic input resolution, and enhanced chain-of-thought reasoning to improve robustness for spatio-temporal grounding.

\section*{C. Prompt Details for Closed-Source Model}

\begin{figure*}[ht]
    \centering
    \includegraphics[width=0.7\linewidth]{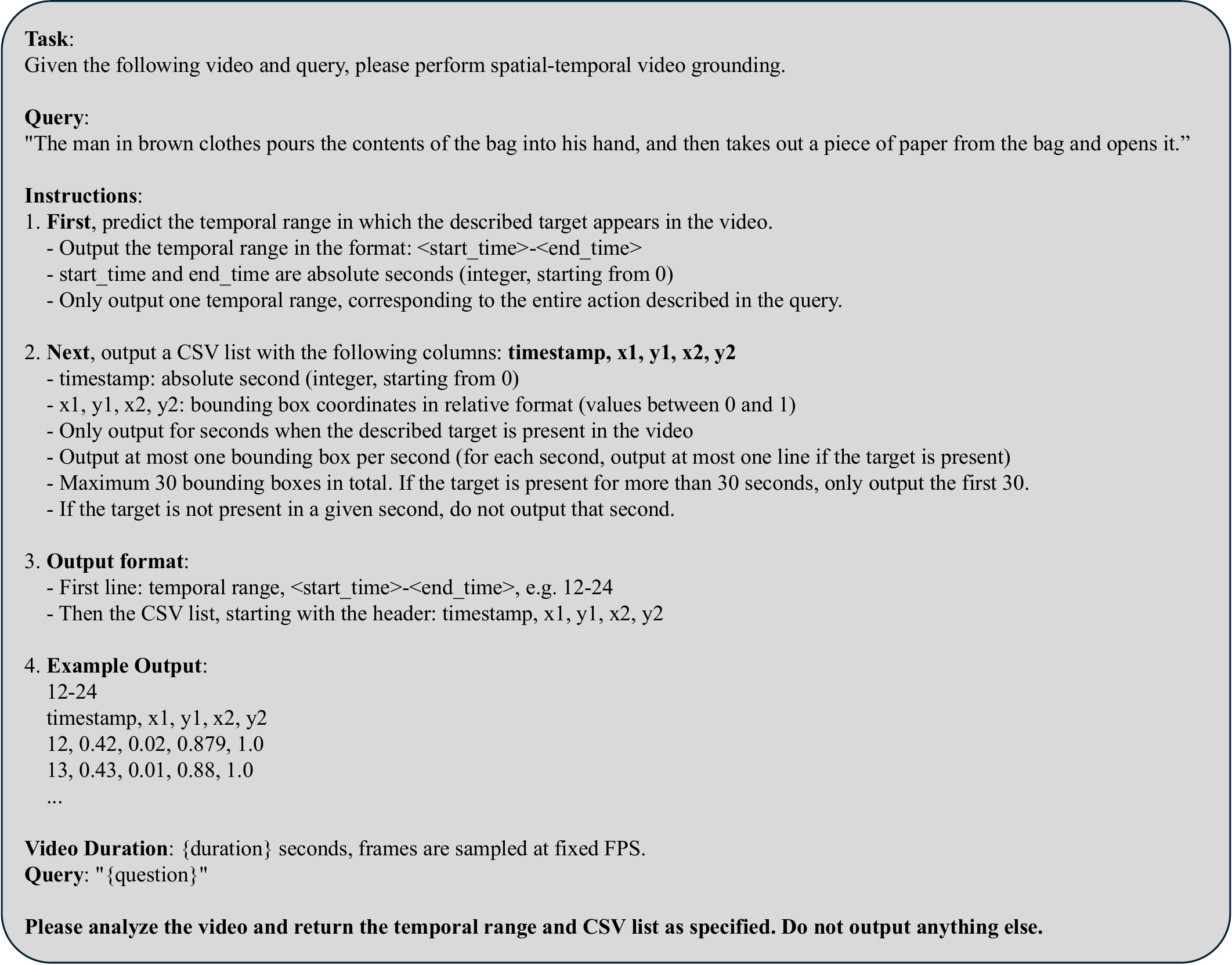}
    \caption{Prompt design for STVG with closed-source models, specifying temporal and spatial output formats.}
    \label{fig:prompt}
\end{figure*}

To investigate the STVG performance of closed-source models, we also experiment with the APIs of closed-source models, including GPT-4o and Gemini-2.5 Pro. To ensure these models generate outputs in the required format, we designed the prompt shown in Fig.~\ref{fig:prompt}.

\end{document}